\documentclass[10pt,twocolumn,letterpaper]{article}

\usepackage[pagenumbers]{cvpr} 

\usepackage{graphicx}
\usepackage{amsmath}
\usepackage{amssymb}
\usepackage{booktabs}
\usepackage{adjustbox}
\usepackage{multirow}
\usepackage{dsfont}
\usepackage{color, colortbl}
\usepackage{xspace}
\usepackage{float}
\usepackage{enumitem}
\usepackage{pifont}
\usepackage[accsupp]{axessibility}

\usepackage[pagebackref,breaklinks,colorlinks]{hyperref}

\usepackage[capitalize]{cleveref}
\Crefname{section}{Section}{Sections}
\crefname{section}{Sec.}{Secs.}
\Crefname{align}{Equation}{Equations}
\crefname{align}{Eq.}{Eqs.}
\Crefname{equation}{Equation}{Equations}
\crefname{equation}{Eq.}{Eqs.}
\Crefname{figure}{Figure}{Figures}
\crefname{figure}{Fig.}{Figs.}
\Crefname{table}{Table}{Tables}
\crefname{table}{Tab.}{Tabs.}

\newcommand\minisection[1]{\vspace{1mm}\noindent \textbf{#1}}
\newcommand{\overbar}[1]{\,\overline{\!{#1}}}
\newcommand{\model}{SaFT\xspace}
\newcommand{\cmark}{\ding{51}}

\definecolor{Gray}{gray}{0.9}

\def\cvprPaperID{2246} 
\def\confName{CVPR}
\def\confYear{2022}

\begin{document}

\title{Semantic-aligned Fusion Transformer for One-shot Object Detection}

\author{
Yizhou Zhao\thanks{The work was done when the author was with MSRA as an intern.} $^1\quad\quad$  Xun Guo$^2\quad\quad$  Yan Lu$^2$ \\
$^1$Carnegie Mellon University  $\qquad^2$Microsoft Research Asia \\
\small \texttt{yizhouz@andrew.cmu.edu} $\qquad$ \texttt{$\{$xunguo, yanlu$\}$@microsoft.com} \\
}
\maketitle

\begin{abstract}
One-shot object detection aims at detecting novel objects according to merely one given instance. With extreme data scarcity, current approaches explore various feature fusions to obtain directly transferable meta-knowledge. Yet, their performances are often unsatisfactory. In this paper, we attribute this to inappropriate correlation methods that misalign query-support semantics by overlooking spatial structures and scale variances. Upon analysis, we leverage the attention mechanism and propose a simple but effective architecture named Semantic-aligned Fusion Transformer (\model) to resolve these issues. Specifically, we equip \model with a vertical fusion module (VFM) for cross-scale semantic enhancement and a horizontal fusion module (HFM) for cross-sample feature fusion. Together, they broaden the vision for each feature point from the support to a whole augmented feature pyramid from the query, facilitating semantic-aligned associations. Extensive experiments on multiple benchmarks demonstrate the superiority of our framework. Without fine-tuning on novel classes, it brings significant performance gains to one-stage baselines, lifting state-of-the-art results to a higher level.
\end{abstract}

\vspace{-0.5cm}
\section{Introduction}
\label{sec:intro}
Recent years have witnessed the flourish of large-scale perception systems like \cite{brown2020gpt, jumper2021alphafold}. Yet it has a long way to go towards real human-like intelligence. Being one of the underlying problems, few-shot learning received more and more interest from language \cite{yu2018diverse, geng2019introduction, bao2020distributional, schick2021exploiting} to vision \cite{vinyals2016matching, snell2017protonet, sung2018learning, liu2021goal, kang2019fsodw, fan2020fsod, wang2020frustratingly} related tasks. This scenario aims at learning a well-generalized model with scarcely labeled data, which challenges conventional learning paradigms. 

\begin{figure}[t]
    \begin{center}
        \includegraphics[width=1.0\linewidth]{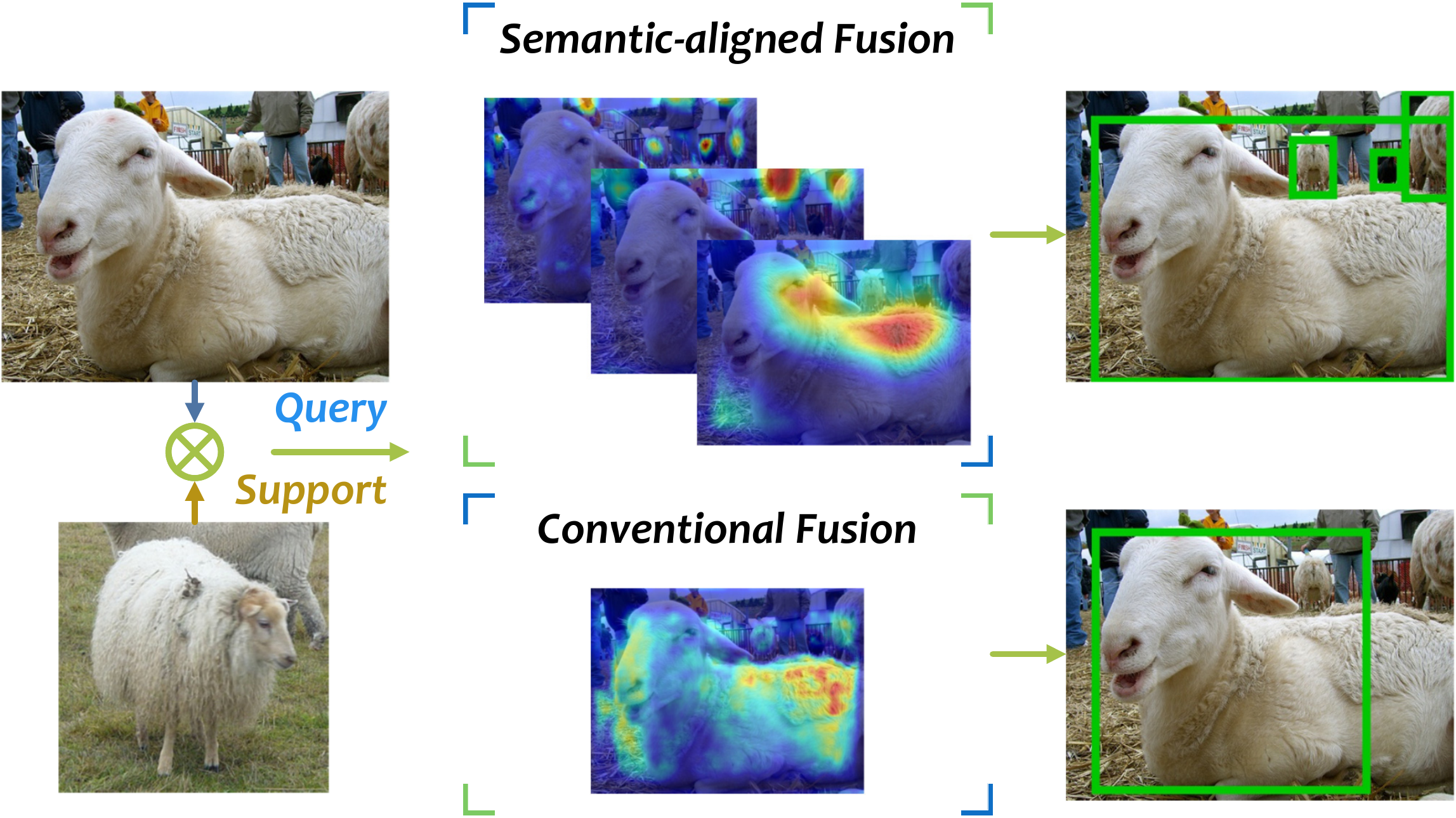}
    \end{center}
    \vspace{-0.5cm}
    \caption{\textbf{Comparison of semantic-aligned fusion and conventional fusion.} Heatmaps and detection results of these two are based on our \model and a baseline with original cross-sample attention accordingly. Comparing the two schemes, semantic-aligned fusion activates more concentrated heatmaps on various feature levels and produces better OSD results.}
    \vspace{-0.5cm}
    \label{fig:comparison}
\end{figure}

To bridge the aforementioned gap in few-shot object detection (FSD), existing literature suggests drawing support from transfer-learning \cite{chen2018lstd, wang2020frustratingly, wu2020mpsr, sun2021fsce, fan2021generalized, zhu2021semantic} or meta-learning \cite{yan2019meta, kang2019fsodw, fan2020fsod, xiao2020wild, zhang2021accurate, li2021beyond, li2021transformation}. Although the former is simple to conduct through pretraining on massive base classes and fine-tuning on scant novel ones, it suffers from the two-stage redundant procedures. The network should always utilize new-coming few-shot data to optimize parameters before it can well recognize these novel classes, thereby limiting its application. In contrast, the latter trend considers meta-knowledge extraction from sampled meta-tasks. This line of frameworks is expected to adapt directly to similarly organized tasks even without online fine-tuning, though it usually helps in performance. At present, this offline meta-learning paradigm is preferred by one-shot object detection (OSD) specific pipelines, with an out-of-the-box availability.

\begin{figure*}[t]
    \begin{center}
        \includegraphics[width=0.8\linewidth]{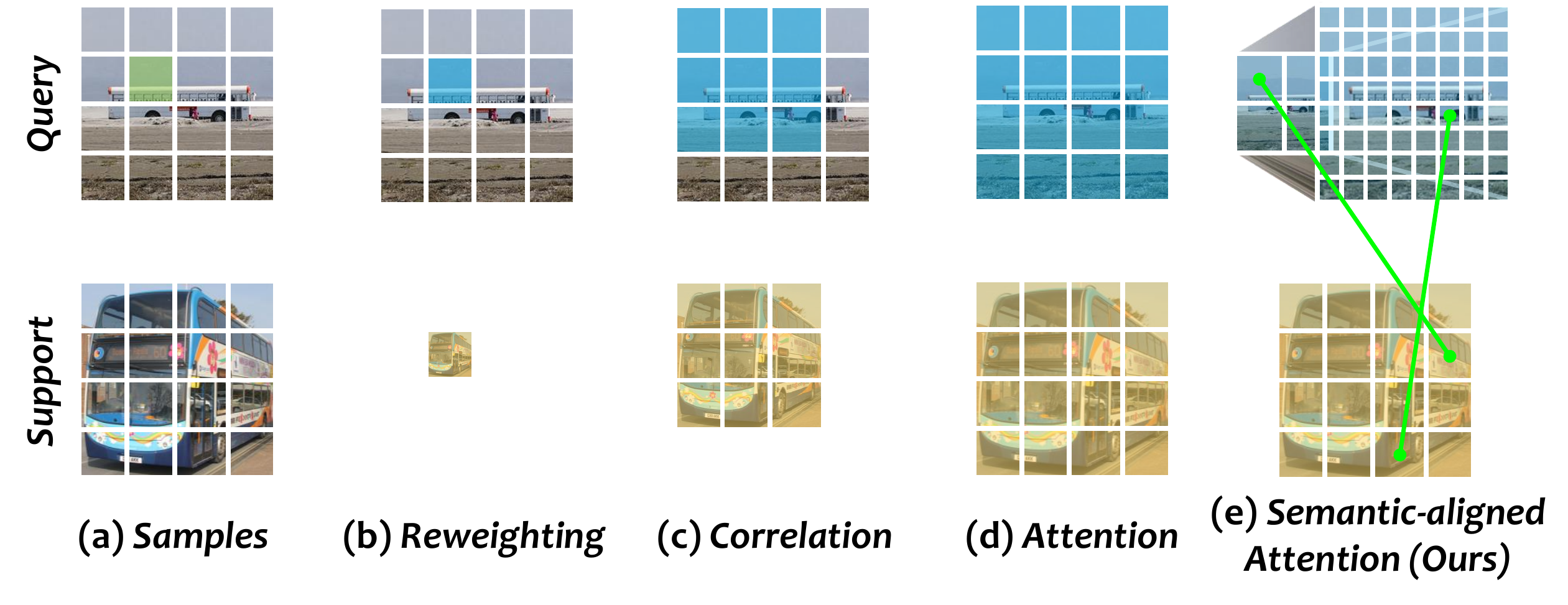}
    \end{center}
    \vspace{-0.6cm}
    \caption{\textbf{Visualization of different fusion approaches.} We present previous fusion schemes in (b), (c), (d), and our proposed semantic-aligned attention as a sort of semantic-aligned fusion in (e). Images are split into patches for illustration, with each patch representing the receptive field of a feature point. The only green patch indicates the query where a response is expected, blue patches are values that contribute to one feature point of the fusion result, and yellow ones are keys that interact with these values. Green lines in (e) suggest two pairs of ideal matches, where different granularities of query features are used. Support samples are scaled in reweighting and correlation schemes to visualize their compression in spatial information.}
    \vspace{-0.4cm}
    \label{fig:fusion}
\end{figure*}

In such a setting, the model should be well constructed to learn the relatedness between a given scene, i.e., the query, and an example patch, i.e., the support. To facilitate this, a series of works \cite{kang2019fsodw, yan2019meta, hsieh2019one, fan2020fsod, perez2020incremental, zhang2021accurate, li2021beyond, osokin2020os2d} investigate cross-sample feature fusion, which augments query features with support representations via sample or ROI level correlation. However, neglecting semantic mismatches in space and scale limits their performances in one-shot scenarios.

Concretely, the traditional paradigm suggests generating a prototype \cite{kang2019fsodw, fan2020fsod} or a kernel \cite{zhang2021accurate} from the support to associate with query features. With most spatial information compressed, the long-range structure-dependent relation in between remains hardly mined. Although pooled prototypes in \cref{fig:fusion}(b) and learned kernels in \cref{fig:fusion}(c) are effective in distinguishing one category from another, they contain fewer positioning priors and thus hinder their localization ability. Furthermore, these schemes match global support representations with local query contexts, regardless of semantic misalignment. An emerging trend \cite{chen2021adaptive, hu2021dense} seeks help from the attention mechanism for adaptive feature fusion. While easing problems discussed to some extent, they commonly focus on feature pairs on a single scale as shown in \cref{fig:fusion}(d), leaving the multi-scale detection task for later anchor-based detector heads. Therefore, it makes no sense when targets are scattered on different scales. For instance, in \cref{fig:fusion}(e), ideal matches for the bus wheel and rear windows lie in two distinct levels of query features, making any single-scale attempt sub-optimal. A simple multi-scale implementation cannot resolve it either, since it fuses the query and the support one scale at a time. Without cross-scale long-range interactions, this rigid manner is likely to fail in cases with semantic missing, such as occlusions or query-support inconsistencies in shape and size.

To encourage more appropriate and sufficient feature interactions in OSD, we propose to adaptively fuse each feature point from the support with each from the query feature pyramid. Thus the original attention mechanism is extended to semantic-aligned attention as illustrated in \cref{fig:fusion}(e). Features from each side are first deconstructed into semantic units, i.e., feature points. Then these units interacts one another in a global manner, not only between query-support sample pairs (horizontally) but also among different scales (vertically). Since objects and parts of objects might exist in different scales and locations, the association process weighted collocates multiple semantic units to make a proper match. In this way, semantic-aligned attention enriches the semantic space that each feature point can utilize, thereby promoting better alignments between the query and the support.

Our Semantic-aligned Fusion Transformer (\model) implements this fusion scheme, with \cref{fig:arch} demonstrating its overall structure. It follows a one-stage proposal-free design and can be easily extended to two-stage pipelines through cascading proposal-based heads. Compared with allied frameworks that employ reweighting or correlation, \model alternatively contains a vertical fusion module (VFM) and a horizontal fusion module (HFM). The former is placed after the feature extractor to together form a Siamese backbone followed by the latter. VFM prepares scale-attended features via vertical attention (VA) in \cref{fig:va}, and HFM utilizes them from query and support with horizontal attention (HA) in \cref{fig:ha}. Note that a single level of support feature interacts with multiple from the other side for a comprehensive view. Thanks to the cross-scale and cross-sample relatedness modeled by the attention mechanism, \model achieves remarkable performance gains in both PASCAL-VOC and MS-COCO datasets.

We conclude our contributions as three-fold.
\begin{enumerate}[leftmargin=12pt,topsep=6pt,parsep=-6pt]
    \item To the best of our knowledge, our Semantic-aligned Fusion Transformer is the first to carry out the offline one-shot object detection task with proposal-free one-stage detectors, producing better performance than state-of-the-art two-stage models.
    \item We discuss the problems of query-support feature fusion and propose a unified attention mechanism to tackle semantic misalignment in space and scale. Our implementation of this can be used as a general fusion neck.
    \item Through qualitative and quantitative experiments, we prove that our novel semantic-aligned fusion is superior to conventional association methods by involving cross-scale long-range relations and collecting more comprehensive meta-knowledge.
\end{enumerate}

\section{Related Work}
\label{sec:related}

\subsection{General Object Detection}
\label{sub:god}
Given a plain image, general object detection aims to localize and classify the objects concerned. Modern detectors can be roughly divided into two categories, namely two-stage proposal-based methods and one-stage proposal-free ones. Two-stage pipelines \cite{ren2015faster, he2017mask, cai2018cascade, he2015spatial, lin2017feature, dai2017deformable} generate a set of class-agnostic region proposals in the first stage and refine as well as classify them into final results in the second. In contrast, one-stage approaches use a class-aware locator to omit the second stage, mostly based on densely placed anchor boxes \cite{redmon2016you, liu2016ssd, lin2017focal} or anchor points \cite{law2018cornernet, duan2019centernet, tian2019fcos, zhang2020bridging}. Different from these, another line of work soaring recently leads a new trend of heuristic-free design. By introducing the attention mechanism, the DETR series \cite{carion2020end, zhu2021deformable, dai2021up} have achieved better performance while being fully end-to-end. Our model builds on one-stage detector FCOS \cite{tian2019fcos} for simplicity, while is rather plug-and-play as a fusion neck.

\subsection{One/Few-shot Object Detection}
\label{sub:fsd}
With sufficient data of base classes while limited samples of novel ones, few-shot scenarios bring more challenges to object detection. Recent work leads two mainstreams in addressing this problem, using transfer-learning or meta-learning techniques. Transfer-learning-based methods \cite{chen2018lstd, zhang2021pnpdet, wang2020frustratingly, wu2020mpsr, qiao2021defrcn, sun2021fsce, fan2021generalized, zhang2021hallucination, zhu2021semantic} follow a two-stage training schema, which is pretraining and fine-tuning, to transfer knowledge from base classes to novel classes. By comparison, the latter trend \cite{yan2019meta, kang2019fsodw, fan2020fsod, xiao2020wild, zhang2021accurate, li2021beyond, hu2021dense, li2021fscn, li2021transformation} recasts the problem in a meta-learning form, encouraging efficient knowledge adaptation through meta-tasks sampling and region-based metric-learning. OSD is an extreme case of FSD with merely one label available for each class to detect. Less data needs more generalization, spawning a series of offline models \cite{hsieh2019one, chen2021adaptive} further explore similarity metrics and abandon the fine-tuning phase. While with different task settings, these methods share a common regional similarity comparison strategy with most FSD networks adopting metric-learning \cite{yan2019meta, karlinsky2019repmet, xiao2020wild, sun2021fsce, li2021transformation}. In other words, they highly rely on region proposals, which can be unpredictable in low-shot scenarios. Different from above, our approach learns the metric in a proposal-free manner, facilitating higher efficiency and flexibility.

\subsection{Multi-scale Feature Fusion}
\label{sub:msf}
Unlike humans that are born with a continuously zooming field of vision, modern convolutional feature extractors usually down-sample images in a discrete way. To mitigate this, multi-scale feature fusion techniques are developed in detection networks, bringing remarkable performance boosts. Three paths in the feature pyramid are exploited, i.e., top-down \cite{lin2017feature}, bottom-up \cite{liu2018path} and within-scale \cite{wang2018non, carion2020end}. Recent work further enriches multi-level information interaction through dense as well as various aggregations \cite{lin2019zigzagnet} and the attention mechanism \cite{zhang2020feature, zhao2021graphfpn}. Although cross-sample feature fusion is widely investigated in one/few-shot problems \cite{vinyals2016matching, sung2018learning, hsieh2019one, doersch2020crosstransformers, xiao2020wild, osokin2020os2d, chen2021adaptive, lin2021cat, perrett2021temporal}, its cross-scale counterpart is relatively scarce. Hence, we consider aggregating these two dimensions and propose a unified attention mechanism for feature fusion between samples and among scales. Compared with its counterparts, this design experimentally helps in semantic alignments.

\section{Method}
\label{sec:method}

\begin{figure*}[t]
    \begin{center}
        \includegraphics[width=1.0\linewidth]{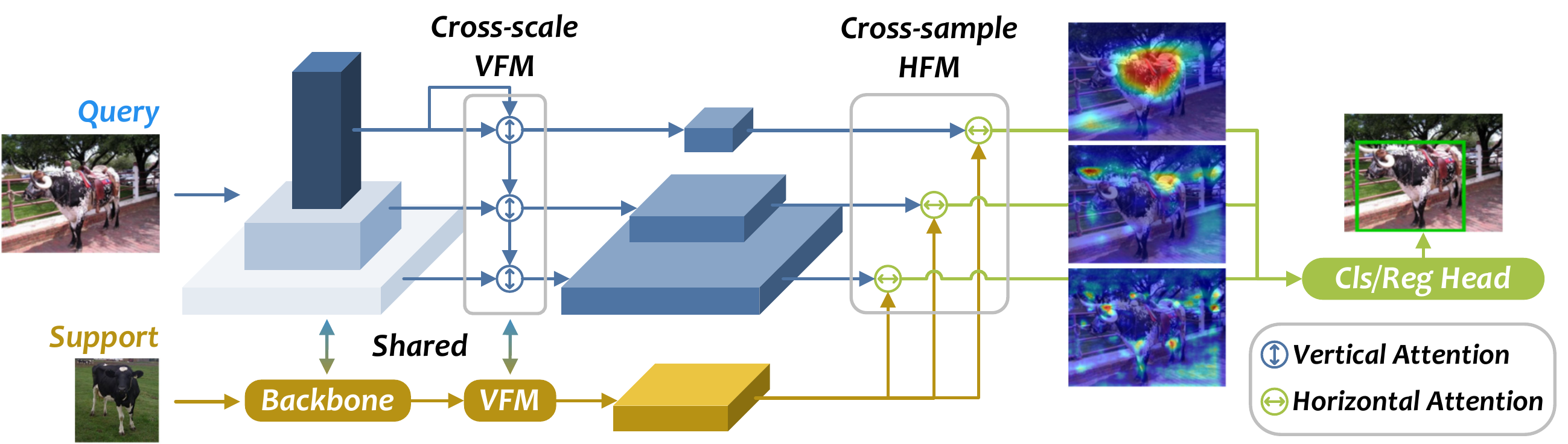}
    \end{center}
    \vspace{-0.4cm}
    \caption{\textbf{The architecture of Semantic-aligned Fusion Transformer for One-shot Object Detection.} Darker color indicates features from deeper layers in backbone, the same in \cref{fig:va}. VFM and HFM are vertical fusion module and horizontal fusion module separately.}
    \vspace{-0.4cm}
    \label{fig:arch}
\end{figure*}

\subsection{Problem Definition}
\label{sub:def}
As in previous literature \cite{hsieh2019one, chen2021adaptive}, the one-shot object detection task constitutes of two sets of instances $\mathcal{D}=\mathcal{D}_\textit{base}\cup\mathcal{D}_\textit{novel}$, where $\mathcal{D}_\textit{base}$ denotes a large base set with numerous available annotations and $\mathcal{D}_\textit{novel}$ stands for a small novel set including only one instance per category. Note that the base classes $\mathcal{C}_\textit{base}$ in $\mathcal{D}_\textit{base}$ and novel classes $\mathcal{C}_\textit{novel}$ in $\mathcal{D}_\textit{novel}$ are mutually exclusive, i.e., $\mathcal{C}_\textit{base}\cap\mathcal{C}_\textit{novel}=\varnothing$.

We consider this problem in a meta-learning fashion akin to \cite{kang2019fsodw, fan2020fsod, li2021beyond} whilst omitting the fine-tuning phase to constrain the setting to fully offline like \cite{hsieh2019one, chen2021adaptive}. Given a query image $Q$ and a support patch $S$, the task is to find all instances of the same category as $S$ with their bounding boxes in $Q$. The base set $\mathcal{D}_\textit{base}$ is provided in training to generate both queries $Q_\textit{base}$ and supports $S_\textit{base}$ while the novel set $\mathcal{D}_\textit{novel}$ is utilized in testing for supports $S_\textit{novel}$ only.

\subsection{Framework}
\label{sub:fw}
We propose a concise framework, termed Semantic-aligned Fusion Transformer (\model), to settle our motivation. The overall architecture is sketched in \cref{fig:arch}. It adopts a Siamese backbone for aligned query-support feature extraction, a shared vertical fusion module (VFM) to enrich per-sample semantic hierarchically, and a subsequent horizontal fusion module (HFM) to aggregate information from both samples for later classification and regression.

\subsection{Feature Fusion via Dense Attention}
\label{sub:da}
Initially introduced in natural language processing \cite{vaswani2017attention} and then borrowed to vision tasks\cite{wang2018non, carion2020end, dosovitskiy2021an}, attention mechanism is known by its inductive bias in modeling long-range information. More specifically in location-aware tasks like detection, positional encodings \cite{parmar2018image, bello2019attention, carion2020end} are adopted upon multi-head attention (MHA) to promote a permutation-variant architecture with
\begin{align}
    \label{eq:pma}
    \text{PMA}(Q,K,V)=\text{MHA}(Q+\text{P}(Q),K+\text{P}(K),V)
\end{align}
where PMA abbreviates position-encoded multi-head attention and P represents positional encoding.

Based on \cref{eq:pma}, we express our dense attention (DA) as follows
\begin{align}
    \label{eq:da}
    \text{DA}(F^Q, F^K) = \text{LN}(F^Q + \text{PMA}(F^Q, F^K, F^K))
\end{align}
where $F^Q \in \mathbb{R}^{h_Q w_Q \times d_Q}$, $F^K \in \mathbb{R}^{h_K w_K \times d_K}$, and $\text{LN}$ denotes layer normalization. Akin to the decoder in \cite{carion2020end} which captures a dense relationship between encoded features and object queries to decode, DA is expected to model a point-to-point correlation between $F^Q$ and $F^K$, thus named as it is. We further extends DA in form of self-attention (SA)
\begin{align}
    \label{eq:sa}
    \text{SA}(F^Q) = \text{DA}(F^Q, F^Q)
\end{align}
and cross-attention (CA)
\begin{align}
    \text{CA}(F^Q, F^K) &= \text{DA}(F^Q, F^K) = F^{K \rightarrow Q}
    \\
    \label{eq:ca}
    \text{CAF}(F^Q, F^K) &= \text{LN}(F^{K \rightarrow Q} + \text{FFN}(F^{K \rightarrow Q}))
\end{align}
where CAF indicates CA with a consecutive feed-forward network (FFN) and an add-and-norm.

Upon these, we propose two sorts of attention blocks, horizontal attention and vertical attention. The elemental procedure of both of them is consistent, with an SA before a CA. This design helps with adaptiveness since SA selectively expresses information from the query side, and CA weighted balances the two sides.

\subsection{Cross-sample Horizontal Attention}
\label{sub:ha}

DA-based cross-sample horizontal attention (HA) is designed for fusion between features from $Q$ and $S$ samples. For comparison, we first concisely review conventional convolution-based approaches in FSD/OSD tasks and then introduce our methods.

Beginning with a pair of features $F^Q$ and $F^S$ extracted from the query and the support, traditional pair-wise operations either extract a prototype or learn a kernel of $S$ as
\begin{align}
    \phi(F^S) = z^S
\end{align}
and then obtain a class-specific enhanced feature with channel-wise multiplication as \cref{fig:fusion}(b) or convolution as \cref{fig:fusion}(c)
\begin{align}
    \tilde{F}^Q = F^Q \odot z^S
\end{align}
where $\tilde{F}^Q$ denotes the enhanced query feature.

This schema highlights class-related information from support samples while discarding most of the spatial semantics. Furthermore, since the class-related $z^S$ represents the whole support patch while its target is a local area from the query, this global-to-local correlation process may lead to a misalignment in space and scale.

\begin{figure}[t]
    \begin{center}
        \includegraphics[width=0.95\linewidth]{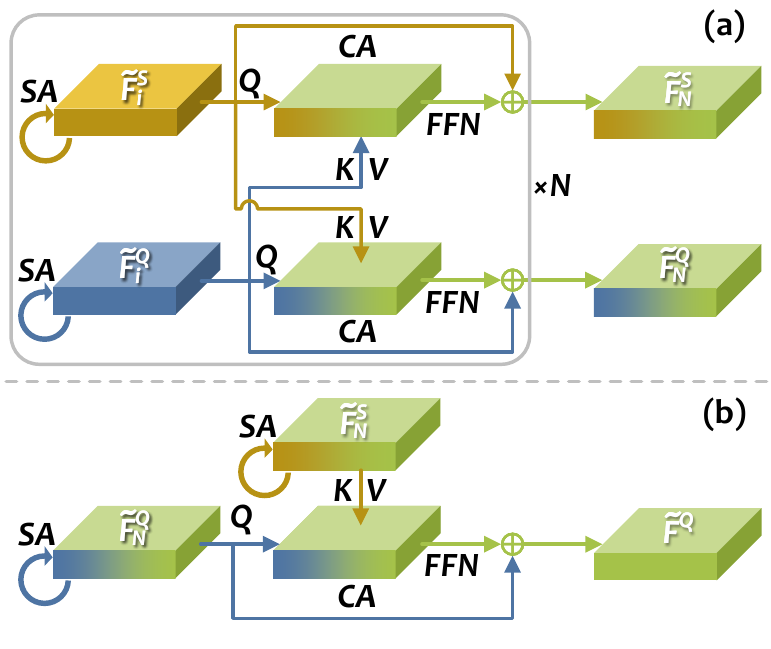}
    \end{center}
    \vspace{-0.7cm}
    \caption{\textbf{The horizontal attention (HA) block.} HA includes two sequential processes. (a) Iterative two-way fusion process with pairs of SA and CAF. (b) Finalizing one-way aggregation process with SAs and a CAF.}
    \vspace{-0.4cm}
    \label{fig:ha}
\end{figure}

In contrast, HA interacts every location pairs of $F^Q$ and $F^S$ as \cref{fig:fusion}(d) shows. A single block of HA presented in \cref{fig:ha} constitutes from a pair of SA and CAF 
\begin{align}
    \overbar{F}^Q_{i+1} &= \text{SA}(\tilde{F}^Q_{i})
    \\
    \overbar{F}^S_{i+1} &= \text{SA}(\tilde{F}^S_{i})
    \\
    \tilde{F}^Q_{i+1} &= \text{CAF}(\overbar{F}^Q_{i+1}, \overbar{F}^S_{i+1})
    \\
    \tilde{F}^S_{i+1} &= \text{CAF}(\overbar{F}^S_{i+1}, \overbar{F}^Q_{i+1})
\end{align}
where overlines and tildes stand for self-attended features and cross-attended features accordingly. For more sufficient feature interactions, these operations conduct iteratively with initial $\tilde{F}^Q_{0}=F^Q$, $\tilde{F}^S_{0}=F^S$ and $i=0,\dots,N-1$. Ending attention layers aggregate features from both sides
\begin{align}
    \label{eq:ha}
    \tilde{F}^Q = \text{HA}(F^Q, F^S) = \text{CAF}(\text{SA}(\tilde{F}^Q_{N}), \text{SA}(\tilde{F}^S_{N}))
\end{align}
We term these chained blocks of HA as a horizontal fusion module (HFM).

Intuitively, HFM conducts global-to-global similarity matching and expression. One by one it associates each feature point from the query and one from the support, regardless of their positions. This schema aligns features from two sides in a deformable and reorganizable manner, thereby making them more comparable.

\subsection{Cross-scale Vertical Attention}
\label{sub:va}
\begin{figure}[t]
    \begin{center}
        \includegraphics[width=0.95\linewidth]{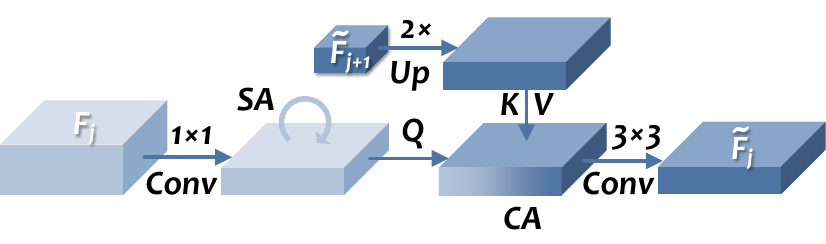}
    \end{center}
    \vspace{-0.6cm}
    \caption{\textbf{The vertical attention (VA) block.} VA inserts an SA and CA pair between lateral and output convolutions of FPN.}
    \vspace{-1.0cm}
    \label{fig:va}
\end{figure}

Besides mutual interactions between $Q$ and $S$ on a single scale, we enhance multi-scale semantics of each sample with cross-scale vertical attention (VA). The whole procedure is shown in \cref{fig:arch} with a close-up illustration in \cref{fig:va}. To show its ability in semantic alignment, we begin with reviewing feature pyramid networks (FPN).

Widely adopted in object detection, FPN is an efficient plug-in to tackle scale variances. Its building blocks can be written as
\begin{align}
    \label{eq:fpn}
    \tilde{F}_j = \text{Conv}_{3 \times 3}(\text{Conv}_{1 \times 1}(F_j) + \text{Up}_{2 \times}(\tilde{F}_{j+1}))
\end{align}
where $F_j$ and $\tilde{F}_j$ are the level-$j$ feature extracted by backbone and the corresponding result after fusion between neighboring scales, separately. We notice that FPN gathers semantics from higher levels to replenish lower levels locally. Notwithstanding its enriched contexts, this in-place scheme falls short in capturing cross-scale long-range information which can be semantically complementary in OSD, e.g., people in a long line are of the same category while having distinct appearance features.

To this end, we introduce VA. Given a feature pyramid $\{F_j | j=3,\dots,M\}$ having strides $\{2^j | j=3,\dots,M\}$ extracted by backbone, VA starts from the self-enhancement of top level $M$ as
\begin{align}
    \tilde{F}_M = \text{Conv}_{3 \times 3}(\text{SA}(\text{Conv}_{1 \times 1}(F_M)))
\end{align}
In a top-down hierarchy, VA adaptively inquires related information globally from upper levels
\begin{align}
    \overbar{F}_j &= \text{SA}(\text{Conv}_{1 \times 1}(F_j))
\end{align}

\begin{align}
\label{eq:va}
\begin{split}
    \tilde{F}_j &= \text{VA}(F_j, \tilde{F}_{j+1}) \\
    &= \text{Conv}_{3 \times 3}(\text{CA}(\overbar{F}_j, \text{Up}_{2 \times}(\tilde{F}_{j+1})))
\end{split}
\end{align}
where $\tilde{F}_j$ is the enhanced feature from level $j$, which is up-sampled before fusion with the next level for alignment. This pyramidal process, named vertical fusion module (VFM), seeks to aggregate multi-scale global semantics.

Compared with FPN in \cref{eq:fpn}, VFM inserts attention layers between lateral and output convolutions. Rather than linearly combining features in the same position, VFM promotes more flexible cross-scale feature interactions and better matches between query-support representations. The query collects and enriches potential target semantics spread in different positions and scales, whereas the support highlights the main target consistent across scales and dims irrelevant backgrounds. Moreover, VFM cooperates with HFM to expand the attentive field of a support feature point from a single layer to multiple scales as in \cref{fig:fusion}(e). With this point-to-pyramid connection, richer semantics and cross-scale long-range correlations are available in matching, thus aiding query-support alignments.

\begin{table*}[t]
\centering
\footnotesize
\setlength{\tabcolsep}{0.4em}
\adjustbox{width=\linewidth}{
    \begin{tabular}{l|ccccccccccccccccc|ccccc}
    \toprule
    \multirow{2}{*}{Method / Set} & \multicolumn{17}{c|}{Base} & \multicolumn{5}{c}{Novel} \\
    & Plant & Sofa & TV & Car & Bottle & Boat & Chair & Person & Bus & Train & Horse & Bike & Dog & Bird & Mbike & Table & Avg. & Cow & Sheep & Cat & Aero & Avg. \\ \midrule
    FSCE$^\ast$ \cite{sun2021fsce} & \textbf{\color{blue}55.0} & 69.6 & \textbf{\color{blue}81.9} & 83.9 & \textbf{\color{blue}71.9} & 65.9 & 45.2 & 45.9 & 83.6 & 85.4 & 86.4 & \textbf{\color{blue}85.1} & 79.2 & 79.5 & \textbf{\color{blue}83.9} & \textbf{\color{blue}73.1} & 73.5 & 72.3 & 67.0 & 53.9 & \textbf{\color{blue}48.0} & 60.3 \\
    DeFRCN$^\ast$ \cite{qiao2021defrcn} & 51.7 & \textbf{\color{blue}74.4} & 78.3 & \textbf{\color{red}87.1} & 70.8 & \textbf{\color{blue}67.6} & \textbf{\color{blue}52.4} & 61.6 & \textbf{\color{blue}85.0} & 85.8 & \textbf{\color{red}87.6} & 83.1 & 82.0 & \textbf{\color{red}83.8} & 82.8 & 64.0 & \textbf{\color{blue}74.9} & 70.7 & 59.1 & 58.8 & 43.0 & 57.9 \\ \midrule
    CoAE \cite{hsieh2019one} & 30.0 & 54.9 & 64.1 & 66.7 & 40.1 & 54.1 & 14.7 & 60.9 & 77.5 & 78.3 & 77.9 & 73.2 & 80.5 & 70.8 & 72.4 & 46.2 & 60.1 & 83.9 & 67.1 & 75.6 & 46.2 & 68.2 \\
    AIT \cite{chen2021adaptive} & 47.7 & 62.7 & 71.9 & 76.1 & 51.8 & 63.5 & 31.5 & \textbf{\color{blue}70.3} & 84.0 & \textbf{\color{blue}87.2} & 81.2 & 80.8 & \textbf{\color{blue}84.5} & 72.2 & 78.7 & 62.8 & 69.2 & \textbf{\color{blue}86.6} & \textbf{\color{blue}74.3} & \textbf{\color{blue}83.7} & 47.7 & \textbf{\color{blue}73.1} \\
    \rowcolor{Gray} \model (Ours) & \textbf{\color{red}59.7} & \textbf{\color{red}81.3} & \textbf{\color{red}82.4} & \textbf{\color{blue}86.9} & \textbf{\color{red}73.0} & \textbf{\color{red}72.0} & \textbf{\color{red}62.3} & \textbf{\color{red}83.7} & \textbf{\color{red}85.9} & \textbf{\color{red}88.1} & \textbf{\color{blue}86.7} & \textbf{\color{red}87.7} & \textbf{\color{red}87.7} & \textbf{\color{blue}83.5} & \textbf{\color{red}86.1} & \textbf{\color{red}75.1} & \textbf{\color{red}80.1} & \textbf{\color{red}88.1} & \textbf{\color{red}77.0} & \textbf{\color{red}84.3} & \textbf{\color{red}48.5} & \textbf{\color{red}74.5} \\
    \bottomrule
    \end{tabular}
}
\caption{Experimental results on the VOC 2007 test set in terms of AP50 (\%). We evaluate performances of our \model over multiple random runs. \textbf{\color{red} RED}\textbf{/}\textbf{\color{blue} BLUE} indicate SOTA/the second best, the same below. The superscript $^\ast$ indicates results reproduced under the OSD setting.}
\vspace{-0.4cm}
\label{tab:voc}
\end{table*}

\begin{table}[t]
\centering
\footnotesize
\setlength{\tabcolsep}{0.4em}
\adjustbox{width=\linewidth}{
    \begin{tabular}{l|ccccc|ccccc}
    \toprule
    \multirow{2}{*}{Method / Split} & \multicolumn{5}{c|}{Base} & \multicolumn{5}{c}{Novel} \\
     & 1 & 2 & 3 & 4 & Avg. & 1 & 2 & 3 & 4 & Avg. \\ \midrule
    CoAE \cite{hsieh2019one} & 42.2 & 40.2 & 39.9 & 41.3 & 40.9 & 23.4 & 23.6 & 20.5 & 20.4 & 22.0 \\
    AIT \cite{chen2021adaptive} & \textbf{\color{red}50.1} & \textbf{\color{red}47.2} & \textbf{\color{blue}45.8} & \textbf{\color{blue}46.9} & \textbf{\color{blue}47.5} & \textbf{\color{blue}26.0} & \textbf{\color{blue}26.4} & \textbf{\color{red}22.3} & \textbf{\color{blue}22.6} & \textbf{\color{blue}24.3} \\
    \rowcolor{Gray} \model (Ours) & \textbf{\color{blue}49.2} & \textbf{\color{red}47.2} & \textbf{\color{red}47.9} & \textbf{\color{red}49.0} & \textbf{\color{red}48.3} & \textbf{\color{red}27.8} & \textbf{\color{red}27.6} & \textbf{\color{blue}21.0} & \textbf{\color{red}23.0} & \textbf{\color{red}24.9} \\ \bottomrule
    \end{tabular}
}
\caption{Experimental results on the COCO 2017 val set in terms of AP50 (\%). Our results are averaged over multiple runs.}
\vspace{-0.6cm}
\label{tab:coco}
\end{table}

\section{Experiments}
\label{sec:exp}

\subsection{Experimental Setting}
\label{sub:set}

\minisection{Benchmarks.}
We follow the previous work \cite{hsieh2019one, chen2021adaptive} to train and evaluate our model on PASCAL-VOC \cite{everingham2010pascal, everingham2015pascal} and MS-COCO \cite{lin2014microsoft} with the same data splits. For VOC, a split of 20 classes partitions the whole dataset into 16 base classes and 4 novel classes. As for COCO, we divide the whole dataset of 80 classes into four groups, each having 20 classes. By turns, three groups are selected from the four as base classes for training and the remained 20 classes for evaluation. The original setting randomly samples query-support image pairs, which generates a different support patch each time a query image is given, in both training and testing. Differently, we keep the former but replace the latter with fixed seeds as in \cite{kang2019fsodw, wang2020frustratingly, xiao2020wild}. This strategy generates one random patch for each support class from COCO17 val and thus restricts the model to see merely one shot in testing rather than the whole set. Compared with the former evaluation, which is at risk of paralleling query and support with identical images, our proposed setting is closer to practical scenarios and de facto one-shot object detection.

\minisection{Implementation Details.}
Our approach employs FCOS \cite{tian2019fcos} as our base detector with ResNet-101 \cite{he2016resnet} pretrained on ImageNet \cite{russakovsky2015imagenet} as backbone. VFM outputs $\{\tilde{F}_4^Q, \tilde{F}_5^Q, \tilde{F}_6^Q \}$ with strides $\{16, 32, 64\}$ from the query while only the intermediate $\tilde{F}_5^S$ from the support for semantic-aligned fusion. HFM iterates $N=6$ two-way HA blocks. To optimize our network, we use SGD with a mini-batch size of 8, momentum of 0.9 and weight decay of $1e-4$ on both PASCAL-VOC and MS-COCO datasets without online fine-tuning. 

\subsection{Comparison Results}
\label{sub:cmp}

\minisection{PASCAL-VOC.}
We provide performance comparison with current state-of-the-art on VOC in \cref{tab:voc}. The first two rows show results we reproduce with FSD methods under the OSD setting. Our \model consistently outperforms existing approaches, which demonstrates its effectiveness. We achieve around 5.2\% and 1.4\% improvements over the best method in base and novel classes respectively. To be specific, we observe a huge upswing in some categories, e.g., Chair with 9.9\% and Person with 13.4\%. One possible reason might be that objects in these classes have larger diversity in shape and size. Our method aligns query-support semantic units more effectively, thereby favoring these cases. It is also noteworthy that, among all the listed methods, our model is the only one adopting a one-stage framework.

\minisection{MS-COCO.}
Similarly, we report evaluation results of COCO on four different splits in \cref{tab:coco}. In spite of the challenge of COCO, our model achieves 24.9\% novel AP50, better than all existing methods. We further notice that performances on base classes and those on novel classes are not necessarily positively correlated. For example, although \model produces relatively low results on the first split of base classes, it brings a 1.8\% increase on novel classes over the current SOTA, demonstrating the strong generalization of our approach.

\subsection{Ablation Study}
\label{sub:abs}
We investigate into the effectiveness of various components of our proposed \model. Presented in \cref{tab:module,tab:scale}, all relative ablations are conducted on the VOC07 test set with half the batch size as our main experiment and less iterative HA blocks ($N=4$). Single-scale implementations in rows 1-3 of \cref{tab:module} utilize Res-4 feature by default, while the rest multi-scale ones use level-4,5,6 features as \model.

\begin{table}[t]
\centering
\footnotesize
\setlength{\tabcolsep}{0.4em}
\adjustbox{width=0.7\linewidth}
	{\begin{tabular}{l|l|cc}
			\toprule
            Cross-scale & Cross-sample & Base & Novel \\ \midrule
            w/o & Reweighting \cite{kang2019fsodw} & 61.8 & 53.7 \\
            w/o & Correlation \cite{zhang2021accurate} & 64.1 & 54.2 \\
            w/o & HFM & 74.6 & 65.8 \\ \midrule
            FPN \cite{lin2017feature} & Reweighting & 72.3 & 62.9 \\
            FPN & Correlation & 76.6 & 61.6 \\
            FPN & HFM & 79.6 & 69.2 \\ \midrule
            VFM & Reweighting & 72.8 & 64.2 \\
            VFM & Correlation & 77.7 & 64.3 \\
            \rowcolor{Gray} VFM & HFM & 79.5 & 71.7 \\
			\bottomrule
	\end{tabular}}
\caption{Ablation study for different modules of \model on VOC. Cross-scale and cross-sample are feature fusion techniques that interact among scales and between samples accordingly.}
\vspace{-0.4cm}
\label{tab:module}
\end{table}

\minisection{Impact of different modules.}
In \cref{tab:module}, reweighting \cite{kang2019fsodw} and correlation \cite{zhang2021accurate} are borrowed for substitution with HFM as our baseline cross-sample fusion operations. We adopt a $5 \times 5$ kernel for correlation and simply pool it to $1 \times 1$ to produce a reweighting prototype. As for cross-scale fusion, we implement pipelines without VFM in rows 1-3 as our baseline and add FPN in rows 4-6 for comparison. Upon these, we go through three phases to finish our exploration of \model. (1) Employ HFM for cross-sample fusion. Comparing the results in rows 1-3, we find that HFM improves by 10.5\% $\sim$ 12.8\% and 11.6\% $\sim$ 12.1\% respectively on base and novel classes. Similar conclusions can be drawn with respect to rows 4-6, showing the superiority of HFM in modeling cross-sample relatedness. With the help of the attention mechanism, HFM deconstructs support features so that they can be matched with the query deformably rather than always as a whole. In this way, semantics from both sides are more aligned. (2) Use FPN for cross-scale fusion. Before looking into VFM, we first extract multi-scale features with FPN, thereby mitigating the problem of scale variation. Besides the 3.4\% $\sim$ 9.2\% performance boost on novel classes with FPN, we notice an interesting phenomenon. Cross-sample fusion approaches with larger receptive fields in the query gain more in single-scale performances, while FPNs attached upon them improve relatively less. This comes not only from the higher baselines they are on, but also from the aggregation of fine-to-coarse features that potentially increase the receptive field of lower levels. (3) Adopt VFM for semantic-aligned fusion. Comparison between rows 6 and 9 presents a 2.5\% promotion by replacing FPN with VFM, as a result of the more adaptive VA in VFM that collects and supplements semantics globally rather than locally. Baseline results also grow with 1.3\% $\sim$ 2.7\% from rows 4-5 to rows 7-8. Finally, VFM and HFM collaborates in semantic-aligned attention and obtains a 17.5\% $\sim$ 18.0\% soar, proving their effectiveness.

\begin{table}[t]
\centering
\footnotesize
\setlength{\tabcolsep}{0.4em}
\adjustbox{width=1.0\linewidth}{
    \begin{tabular}{l|l|ccc|ccc|cc} \toprule
    \multirow{2}{*}{Cross-scale} & \multirow{2}{*}{Cross-sample} & \multicolumn{3}{c|}{Query} & \multicolumn{3}{c|}{Support} & \multirow{2}{*}{Base} & \multirow{2}{*}{Novel} \\
    &  & 4 & 5 & 6 & 4 & 5 & 6 &  &  \\ \midrule
    FPN\cite{lin2017feature} & Corresponding-scale  & \cmark & \cmark & \cmark & \cmark & \cmark & \cmark & 77.4 & 68.7 \\
    FPN & One-to-all-scale & \cmark & \cmark & \cmark &  & \cmark &  & 79.6 & 69.2 \\ \midrule
    VFM & Corresponding-scale  & \cmark & \cmark & \cmark & \cmark & \cmark & \cmark & 77.5 & 69.9 \\
    \rowcolor{Gray} VFM & One-to-all-scale & \cmark & \cmark & \cmark &  & \cmark &  & 79.5 & 71.7 \\ \bottomrule
    \end{tabular}
}
\caption{Ablation study for query-support fusion scales on VOC. One-to-all-scale means associating a single level of support features with all available query features, whereas corresponding-scale is limited to corresponding levels. All experiments use HFM for cross-sample fusion while with different corresponding rules.}
\vspace{-0.4cm}
\label{tab:scale}
\end{table}

\minisection{Impact of query-support fusion scales.}
We examine the way to utilize different levels of features for fusion in \cref{tab:scale}. To begin with, we propose a simple corresponding-scale strategy that fuses each level of support feature with the same level of query feature. Its results are shown in the first and the third row. While with additional levels of support features participating in query-support fusion, its results turn out relatively lower. This counter-intuitive result might be due to a lack of generality in multi-level support features. Corresponding-scale fusion forces single-level fusion and detection, potentially causing multi-level connections to degrade by over-fitting to each level. In fact, a common down-sample ratio for each corresponding level of query-support feature means always searching targets of the same size. This not only undermines semantic alignments but also confuses multi-scale detection. Differently, since one-to-all-scale fusion matches the support with different sizes of targets on multiple scales, it learns more generalized meta-knowledge. This leads to 0.5\% $\sim$ 1.8\% gains in novel classes after shifting from the corresponding-scale strategy. Also, a comparison between rows 2 and 3 shows that VFM with corresponding-scale query-support fusion outperforms FPN with a one-to-all-scale one. We attribute this to the cross-scale long-range correlation modeled by VFM, which mitigates cross-scale semantic misalignment.

\begin{figure}[t]
    \begin{center}
        \includegraphics[width=1.0\linewidth]{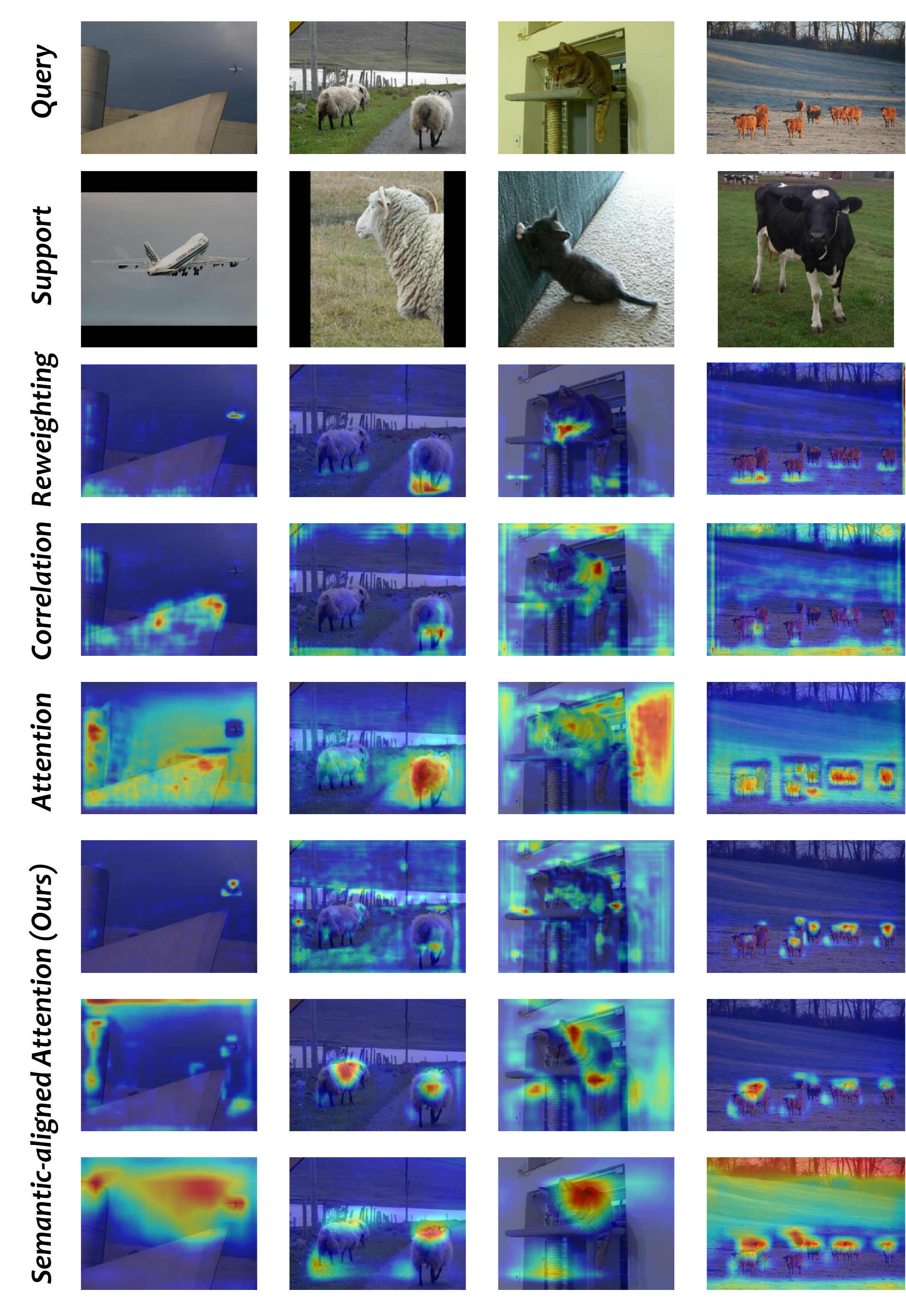}
    \end{center}
    \vspace{-0.65cm}
    \caption{\textbf{Feature map responses of different fusion approaches.} Vertically, the four columns are four different classes in the VOC07 test set. Horizontally, the top two rows are query and support samples for fusion, while the rest are feature map responses based on four fusion paradigms. Rows 3, 4, 5 show visualizations for three single-scale implementations. The last three lines demonstrate results of feature levels 4, 5, and 6 for our \model with VFM and HFM. These four configurations are respectively corresponding to rows 1-3 and 9 in \cref{tab:module}.}
    \vspace{-0.45cm}
    \label{fig:visual}
\end{figure}

\subsection{Qualitative Analysis}
\minisection{Feature map responses of different fusion approaches.}
To explore the behaviors of four fusion paradigms introduced in \cref{fig:fusion}, we visualize their feature map responses in \cref{fig:visual}. These heatmaps are produced by averaging fusion features in all channels. Comparing row 3 with row 4, we can see that convolution-based methods tend to focus on objects that conform to their kernel sizes. Reweighting with a $1 \times 1$ kernel recognizes the tiny airplane in the first column while correlation with a $5 \times 5$ one hardly not. In larger objects, such as the cat in the third column, the heatmap of correlation activates more intensely than that of reweighting. Moreover, most related objects are partially focused due to the query-support mismatch. In contrast, attention-based HFM distributes its attention to more complete object regions. Yet, it activates some unrelated areas as well, especially in cases like the first and third columns, where the query shares a similar background with the support. We attribute the former to scale issues and the latter to the adaptive instinct of attention mechanisms, sometimes misleading. Generally, these conventional techniques are restricted in tackling appearance and scale variations, which leads to their activation biases compared with semantic-aligned attention. Illustrated in the three bottom rows, objects of distinct scales are activated differently in multi-scale heatmaps. As targets grow in size, fusion levels of interest shift from low to high. Clearly shown in the figure, the small-scale airplane, middle-scale sheep, and large-scale cat are highlighted in level 4, 5, and 6 heatmaps, accordingly. As a result, the attention mechanism works in tandem with the semantic-aligned fashion to address feature misalignment as well as scale discrepancies.

\begin{figure}[t]
    \begin{center}
        \includegraphics[width=1.0\linewidth]{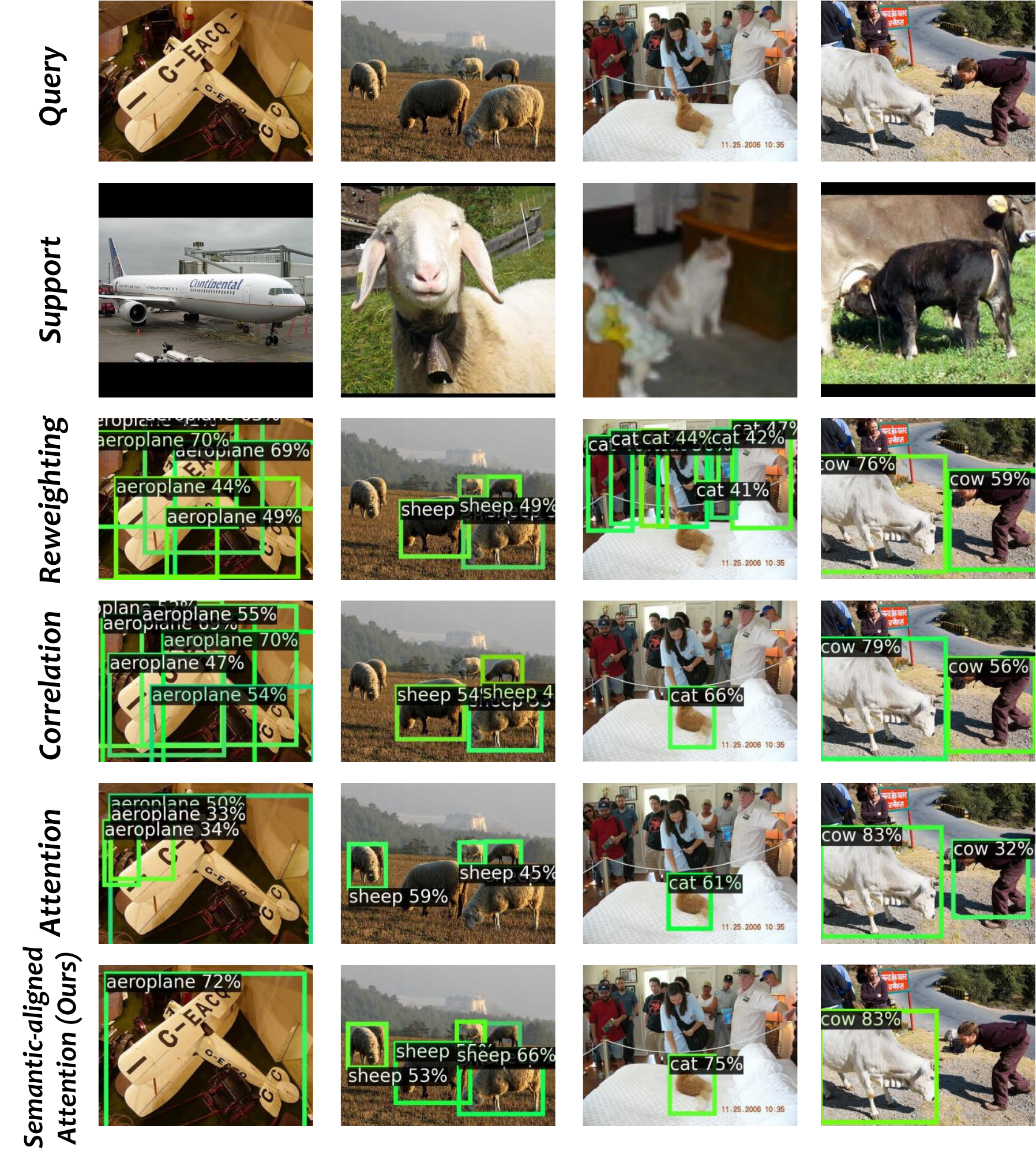}
    \end{center}
    \vspace{-0.6cm}
    \caption{\textbf{Detection results of different fusion approaches.} Four columns show original samples and detection results for each novel class in the VOC07 test set. Results in rows 3-6 are of configurations as rows 1-3 (previous schemes) and 9 (our novel scheme) in \cref{tab:module}.}
    \vspace{-0.6cm}
    \label{fig:detect}
\end{figure}

\minisection{Detection results of different fusion approaches.}
We demonstrate detection results in \cref{fig:detect} for a more intuitive comparison. To be concrete, we explore the figure column by column. The first column clearly shows the differences between convolution-based techniques and attention-based ones. Matching the support patch in local areas of the query, reweighting and correlation are only capable of capturing bounding boxes that cover part of the object. In contrast, attention and semantic-aligned attention produce more accurate results, as they obtain a more global view of the query. Additionally, we can find more false positives in the results of traditional schemes. This is because their misalignment in scale wrongly focuses them on irrelevances. Similar phenomena can be seen in the last column, where a human is recognized as a cow in rows 3-5. From top to bottom, networks successively give higher certainties to the cow and lower to the human. This also proves that space and scale alignments help in query-support associations. In the second column, we examine the ability of different approaches in resolving multi-scale issues. It is worth noticing that rows 3-5 show a downward trend in handling scale variances. Not difficult to understand, the discrepancy between reweighting and correlation is due to the inductive bias of their kernel sizes. With a smaller kernel size, reweighting is better at detecting smaller objects. In terms of attention, it fails to resolve scale variations solely dependent on single-scale matching. Thus, HFM in conjunction with VFM forming semantic-aligned attention presents better predictions. Next, we find an interesting anomaly in the third column. Despite the success of others, reweighting treats all humans as cats while ignoring the real cat. We attribute this to potential semantic conflicts in the support. As in this example, there is a human baby and a cat in the support patch. Then the compression of spatial structure is likely to distract detectors from the never-before-seen cat to the more familiar human that lies in base classes. By contrast, preserving structural information with larger convolution kernels or matching deformably with attention mechanisms remedies this.

\section{Limitation Discussion}
\label{sec:lim}
As an OSD-specific pipeline, our approach cannot be easily extended to more-shot circumstances. The Siamese design limits its inputs to a paired form, necessitating the construction of a dedicated feature extractor and aggregator for multiple support instances. Also, we notice that our model needs a small learning rate and a long schedule to converge. Hence, we adopt a learning rate of 0.001 instead of the common 0.02. Larger learning rates may result in instability in training. Future work can improve in these aspects.

\section{Conclusion}
\label{sec:con}
In this paper, we look into the one-shot object detection task from the perspective of space and scale. The proposed Semantic-aligned Fusion Transformer substantially alleviates the underlying query-support misalignment in traditional feature fusion techniques employed by existing schemes. Despite its intuitiveness, our model achieves state-of-the-art performances on various benchmarks.

\appendix
\section*{\centering Supplementary Material}

\section{Implementation Details}
\label{sub:loss}
Since \model is a general fusion neck, it may employ any detector-specific loss. For the FCOS-based detector as in our main paper, we simply combine losses of FCOS \cite{tian2019fcos} and DETR \cite{carion2020end}. Further implementations on two-stage detectors can try correspondingly related losses as well as metric-learning ones.

Given network predictions for classification $c$, regression $r$, center-ness $t$ as defined in \cite{tian2019fcos} and their corresponding targets $c^*$, $r^*$, $t^*$ respectively, we present our loss function as follows
\begin{equation}
	\begin{split}
		\mathcal{L} &= \frac{\lambda_\textit{cls}}{N}\sum_{x,y}{\mathcal{L}_\textit{cls}(c_{x,y}, c_{x,y}^*)}
		\\
		&+ \frac{\lambda_\textit{reg}}{N_\textit{pos}}\sum_{x,y}{\mathds{1}_{\{c_{x,y}^*>0\}}\mathcal{L}_\textit{reg}(r_{x,y}, r_{x,y}^*)}
		\\
		&+ \frac{\lambda_\textit{ctn}}{N_\textit{pos}}\sum_{x,y}{\mathds{1}_{\{c_{x,y}^*>0\}}t_{x,y}^*\mathcal{L}_\textit{ctn}(t_{x,y}, t_{x,y}^*)}
	\end{split}
\end{equation}
where $\mathcal{L}_\textit{cls}$ is focal loss \cite{lin2017focal}, $\mathcal{L}_\textit{reg}$ is a joint of L1 loss and GIoU loss \cite{rezatofighi2019generalized} as in \cite{carion2020end}, and $\mathcal{L}_\textit{ctn}$ is binary cross entropy (BCE) loss for center-ness as in \cite{tian2019fcos}. $\lambda_\textit{cls}$, $\lambda_\textit{reg}$ and $\lambda_\textit{ctn}$ are balance weights, being 20, 2 and 0.5 so as to keep three terms 
in the same scale. $N$ and $N_\textit{pos}$ denote the number of all locations and that of positive samples.

\section{Ablation Study}
For further insights into \model, we use the same setup as in our main paper to perform more ablation studies. Experiments are carried out on VOC with a mini-batch size of 4. \cref{tab:level} adopts $N=4$ HA blocks by default, and \cref{tab:iter} discusses the effect of HA block numbers.

\begin{table}[t]
	\centering
	\footnotesize
	\setlength{\tabcolsep}{0.4em}
	\adjustbox{width=1.0\linewidth}{
		\begin{tabular}{l|ccc|ccc|cc} \toprule
			\multirow{2}{*}{Cross-sample} & \multicolumn{3}{c|}{Query} & \multicolumn{3}{c|}{Support} & \multirow{2}{*}{Base} & \multirow{2}{*}{Novel} \\
			& 4 & 5 & 6 & 4 & 5 & 6 &  &  \\ \midrule
			\multirow{3}{*}{Corresponding-scale} & \cmark &  &  & \cmark &  &  & 75.8 & 61.4 \\
			& \cmark & \cmark &  & \cmark & \cmark &  & 79.2 & 70.0 \\
			& \cmark & \cmark & \cmark & \cmark & \cmark & \cmark & 77.5 & 69.9 \\ \midrule
			\multirow{5}{*}{One-to-all-scale} & \cmark &  &  &  & \cmark &  & 76.0 & 62.5 \\
			&  & \cmark &  &  & \cmark &  & 76.3 & 60.7 \\
			&  &  & \cmark &  & \cmark &  & 74.1 & 57.7 \\
			& \cmark & \cmark &  &  & \cmark &  & 79.5 & 67.9 \\ 
			&  & \cmark & \cmark &  & \cmark &  & 79.4 & 65.8 \\ \midrule
			\multirow{3}{*}{} & \cmark & \cmark & \cmark & \cmark &  &  & 78.7 & 68.1 \\
			\rowcolor{Gray} One-to-all-scale & \cmark & \cmark & \cmark &  & \cmark &  & 79.5 & 71.7 \\
			& \cmark & \cmark & \cmark &  &  & \cmark & 78.2 & 69.7 \\ \bottomrule
		\end{tabular}
	}
	\caption{Ablation study for feature levels used in fusion on VOC. One-to-all-scale means associating a single level of support features with all available query features, whereas corresponding-scale is limited to corresponding levels. All experiments use VFM for cross-scale fusion and HFM for cross-sample fusion.}
	\vspace{-0.2cm}
	\label{tab:level}
\end{table}

\subsection{Feature Levels Used in Fusion}
In \cref{tab:level}, we explore the effect of feature fusion with different query-support feature levels. For corresponding-scale fusion in rows 1-3, we can see that more levels of features utilized do not necessarily mean better performances. Concretely, adding level 5 to feature fusion provides a huge improvement (8.6\%), while further adding level 6 leads to a slight drop (0.1\%). This is probably because the semantic misalignment in feature fusion at level 6 distracts the detector. Next, rows 4-8 show results of fusion with different levels of query features. From these, we observe a coarser query feature generally benefits the performance, with the one-to-all-scale corresponding strategy. This result is intuitive since coarser feature maps provide stronger positioning priors. In addition, more levels of query features also help, which is different from the corresponding-scale scheme. Then we include all three levels of query features and investigate how support features make a difference. Results in the last three lines show that the level 5 support feature obtains the best performance. We consider this in two folds. On one hand, it is likely due to a preference for this data distribution. On the other hand, as the intermediate one among 4, 5, and 6, this level is relatively more comparable with the whole feature pyramid.

\begin{figure*}[t]
	\begin{center}
		\includegraphics[width=1.0\linewidth]{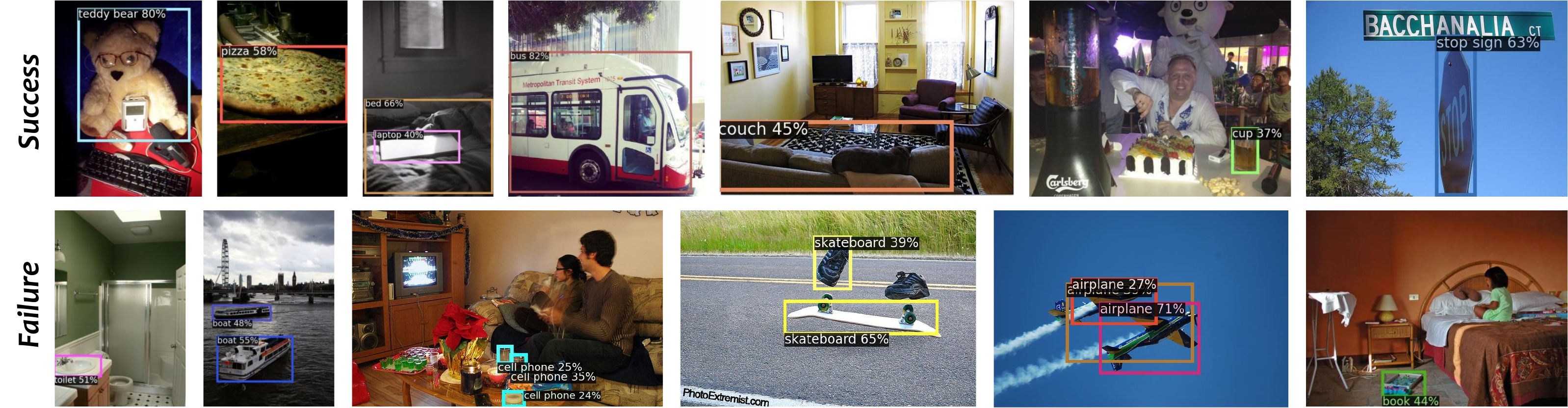}
	\end{center}
	\caption{Success and failure cases of \model on COCO novel classes. We visualize bounding boxes with scores over 0.2.}
	\label{fig:case}
\end{figure*}

\begin{table*}[t]
	\centering
	\footnotesize
	\setlength{\tabcolsep}{0.4em}
	\adjustbox{width=\linewidth}{
		\begin{tabular}{l|ccccccccccccccccc|ccccc}
			\toprule
			\multirow{2}{*}{Base / Set} & \multicolumn{17}{c|}{VOC Base} & \multicolumn{5}{c}{VOC Novel} \\
			& Plant & Sofa & TV & Car & Bottle & Boat & Chair & Person & Bus & Train & Horse & Bike & Dog & Bird & Mbike & Table & Avg. & Cow & Sheep & Cat & Aero & Avg. \\ \midrule
			VOC & 59.7 & 81.3 & 82.4 & 86.9 & 73.0 & 72.0 & 62.3 & 83.7 & 85.9 & 88.1 & 86.7 & 87.7 & 87.7 & 83.5 & 86.1 & 75.1 & 80.1 & 88.1 & 77.0 & 84.3 & 48.5 & 74.5 \\
			COCO & 11.1 & 46.1 & 53.6 & 67.2 & 16.8 & 24.0 & 22.9 & 6.9 & 45.0 & 45.1 & 50.2 & 10.0 & 46.7 & 42.3 & 8.9 & 23.3 & 32.5 & 86.8 & 76.4 & 40.5 & 34.7 & 59.6 \\
			\bottomrule
		\end{tabular}
	}
	\caption{Cross-domain comparison results on the VOC 2007 test set in terms of AP50 (\%). The first column indicates the base training set, where VOC is trained on 16 VOC base classes and COCO on 60 COCO base classes non-overlapping with all VOC categories. Note that the first 16 categories (columns 2-17) are base classes for the VOC base training set while novel classes for COCO.}
	\label{tab:c2v}
\end{table*}

\begin{table}[t]
	\centering
	\footnotesize
	\setlength{\tabcolsep}{0.4em}
	\adjustbox{width=1.0\linewidth}{
		\begin{tabular}{l|c|c|cc} \toprule
			Cross-sample & \# HA Blocks & \# Parameters & Base & Novel \\ \midrule
			Reweighting & 0 & 55.1M & 72.8 & 64.2 \\
			Correlation & 0 & 55.1M & 77.7 & 64.3 \\
			HFM & 2 & 61.6M & 78.5 & 69.5 \\
			HFM & 4 & 67.9M & 79.5 & 71.7 \\
			HFM & 6 & 74.2M & 79.8 & 72.8 \\ \bottomrule
		\end{tabular}
	}
	\caption{Ablation study for the number of iterative HA blocks on VOC. All experiments employ VFM for cross-scale fusion and the one-to-all-scale scheme for cross-sample fusion.}
	\label{tab:iter}
\end{table}

\subsection{Number of Iterative Fusion Blocks}
From the perspective of performance and complexity, we compare our \model with different numbers of HA blocks in \cref{tab:iter}. Reweighting and correlation are presented in the first two rows as baselines with the same number of parameters, since their only difference is non-parameter pooling. Out of their 55.1M parameters, 53.5M are in the backbone, with the same below. We notice that while correlation beats reweighting on base classes by a large margin, their results on novel classes are very close. This indicates an over-fitting tendency for convolution-based methods with large kernels. Comparing these baselines with no HA blocks to HFM with 2 HA blocks, base and novel AP50 improve by 0.8\% $\sim$ 5.7\% and 5.2\% $\sim$ 5.3\% respectively. These improvements demonstrate the effectiveness of attention-based HFM, with 6.5M more parameters. Also, from rows 3-5, performances on base classes grow by 1.3\% and that on novel classes by 3.3\%. This suggests more HA blocks provide more sufficient fusion, which leads to better results. But as the complexity increases linearly, performance growth gradually slows down.

\section{Comparison Results}
We evaluate cross-domain OSD performances following \cite{kang2019fsodw, wu2020mpsr, qiao2021defrcn}, which selects 60 categories in COCO14 non-overlapping with VOC as base classes and all 20 categories in VOC as novel classes. Other experimental settings are the same as in our main experiments. In the bottom line of \cref{tab:c2v}, performances on different categories vary greatly. For instance, the model produces extremely low results for people and motorbikes. We attribute this to the feature extractor, lack of ability to highlight never-before-seen foregrounds. Although most classes experience a downswing compared with same-domain results, we notice that performances on cows and sheep are basically unchanged. Considering there are several base categories of animals, this suggests our offline \model is easier to adapt to a novel data distribution similar to base classes.

\section{Qualitative Analysis}
We provide extra qualitative visualizations of detected novel objects on COCO in \cref{fig:case}. Success cases are presented in the upper row and failure ones in the lower row. The latter include false positives, e.g., the toilet, missing cases, e.g., the boat, and repeat detections, e.g., the airplane.

\clearpage
{\small
\bibliographystyle{ieee_fullname}
\bibliography{egbib}
}

\end{document}